\pdfoutput=1

\documentclass[11pt]{article}

\usepackage[preprint]{acl}

\usepackage{times}
\usepackage{latexsym}
\usepackage{graphicx}
\usepackage{multirow}
\usepackage{amsmath,enumitem}
\usepackage{booktabs}

\usepackage[T1]{fontenc}

\usepackage[utf8]{inputenc}

\usepackage{microtype}

\usepackage{inconsolata}

\usepackage{graphicx}

\usepackage{inconsolata}
\usepackage{mdframed}
\usepackage{makecell}
\usepackage{amssymb}

\usepackage{amsmath}
\usepackage{algorithm}
\usepackage{algpseudocode}

\usepackage{xspace}
\newcommand{\modelname}{BDC\xspace}

\newcommand{\minisection}[1]{\vspace{0pt}\noindent\textbf{#1.}}

\title{Boost, Disentangle, and Customize: \\
A Robust System2-to-System1 Pipeline for Code Generation}

\author{Kounianhua Du$^1$, Hanjing Wang$^1$, Jianxing Liu$^1$, Jizheng Chen$^1$, \\
{\bf Xinyi Dai$^2$, Yasheng Wang$^2$, Ruiming Tang$^2$, Yong Yu$^1$, Jun Wang$^3$, Weinan Zhang$^1$}\\
  $^1$Shanghai Jiao Tong University, $^2$ Huawei Noah’s Ark Lab, $^3$ University College London \\
  Shanghai, China\\
  \texttt{\{kounianhuadu, wnzhang\}@sjtu.edu.cn}
  }

\begin{document}
\maketitle

\begin{abstract}
Large language models (LLMs) have demonstrated remarkable capabilities in various domains, particularly in system 1 tasks, yet the intricacies of their problem-solving mechanisms in system 2 tasks are not sufficiently explored. 
Recent research on System2-to-System1 methods surge, exploring the System 2 reasoning knowledge via inference-time computation and compressing the explored knowledge into System 1 process. In this paper, we focus on code generation, which is a representative System 2 task, and identify two primary challenges: (1) the complex hidden reasoning processes and (2) the heterogeneous data distributions that complicate the exploration and training of robust LLM solvers. To tackle these issues, we propose a novel BDC framework that explores insightful System 2 knowledge of LLMs using a MC-Tree-Of-Agents algorithm with mutual \textbf{B}oosting, \textbf{D}isentangles the heterogeneous training data for composable LoRA-experts, and obtain \textbf{C}ustomized problem solver for each data instance with an input-aware hypernetwork to weight over the LoRA-experts, offering effectiveness, flexibility, and robustness. This framework leverages multiple LLMs through mutual verification and boosting, integrated into a Monte-Carlo Tree Search process enhanced by reflection-based pruning and refinement. Additionally, we introduce the DisenLora algorithm, which clusters heterogeneous data to fine-tune LLMs into composable Lora experts, enabling the adaptive generation of customized problem solvers through an input-aware hypernetwork. Our contributions include the identification of critical challenges in existing methodologies, the development of the MC-Tree-of-Agents algorithm for insightful data collection, and the creation of a robust and flexible solution for code generation. This work lays the groundwork for advancing LLM capabilities in complex reasoning tasks,  offering a novel System2-to-System1 solution.



\end{abstract} 

\section{Introduction}

\begin{figure}[t]
    \centering
    \includegraphics[width=1.0\linewidth]{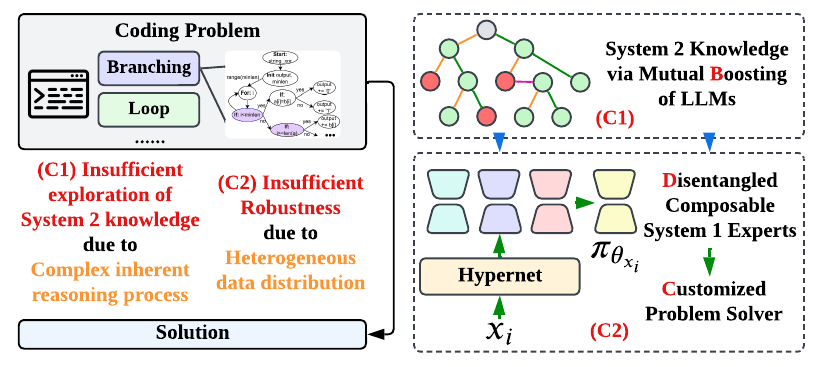}
    \caption{Illustration of the motivation.}
    \label{fig:intro}
\end{figure}

Large language models show significant intelligence in various domains, striking both the academic and industrial institutions. Despite their prominent problem-solving abilities in system 1 tasks, the mechanism behind the system 2 task solving procedure remain opaque. In this paper, we focus on the code generation task, which emerges as a captivating frontier \citep{zheng2023codegeex, roziere2023code, shen2023pangu}, promising to revolutionize software development by enabling machines to write and optimize code with minimal human intervention. Recent research of llms for code focus on inference-time computation (System 2 methods) \citep{yang2024chain, yao2024tree, zhang2023planning} and post-training. While during post-training, distilling system 2 knowledge into system 1 backbones is important and widely-used \citep{yu2024distilling21}. 

However, the complex hidden reasoning process and the heterogeneous data distribution pose challenges to the existing System2-to-System1 pipeline. On one hand, the hidden reasoning process for code generation is complex and hard to explore \textbf{(C1)}. On the other hand, the heterogeneous data distribution, e.g., jumping structure like branching, recursion, etc., makes the existing train-once-for-all strategy hard to fit the complex latent patterns for robust and generalizable llm solvers \textbf{(C2)}. 

For \textbf{(C1)}, we propose to disentangle the problem solving process into problem2thought and thought2solution stages, exploring the inherent reasoning clues via combining the strengths of multiple llms by mutually-verification and boosting. The exploration is integrated into a Monte-Carlo Tree Search process, where reflexion-based pruning and refinement are designed for more efficient and effective reasoning clues search.

For \textbf{(C2)}, we propose to disentangle the heterogeneous data into clusters, finetuning llms capable of different aspects of tasks to obtain the meta LoRA experts hub, and then adaptively generate customized problem solver for each code problem.  Concretely, we design an input-aware hypernetwork to generate rank-wise weights over meta LoRA experts for customized problem solver, offering robustness and flexibility.

The main contributions of our work can be summarized below.
\begin{itemize}
    \item \textbf{Identification of problems and novel BDC framework.} We identify the high-reasoning demand and heterogeneous latent patterns problems that hinders the performance of existing methods and propose a BDC framework that explores insightful inherent reasoning clues via multi-llms boosting, generates meta-LoRA experts via finetuning on disentangled data, and offer customized problem solver with an input-aware hypernet for rank-wise LoRA merging.
    \item \textbf{Novel MC-Tree-of-Agents algorithm for insightful data collection.} We disentangle the System 2 solving process into problem2thought and thought2solution stages, integrating the exploration process into a reflexion-based monte carlo tree search armed with pruning and refinement, enabling mutually verification and boosting of different agents for insightful data collection. 
    \item \textbf{Novel DisenLoRA algorithm that offers customized problem solver for robust code generation.} We disentangle the heterogeneous data distribution into clusters on which meta-LoRA experts are trained, and design an input-aware hypernetwork to weight over the LoRA-experts for customized problem solver, offering robustness and flexibility.
\end{itemize}

\section{Related Work}
\subsection{System 2 Methods in LLMs}
Recent research on large language models for System 2 tasks focus on inference-time computation optimization to stimulate the inherent reasoning ability of LLMs. Few-shot learning methods \cite{wang2022code4struct,madaan2022language} utilize the in-context-learning ability of LLMs for enhanced generation. Retrieval-augmented generation (RAG) approaches \cite{nashid2023retrieval,du2024codegragbridginggapnatural} further introduce domain knowledge into LLMs. 
Techniques such as Chain-of-Thought (CoT) \cite{yang2024chain,jiang2024self,li2023structured}, Tree-of-Thought (ToT) \cite{yao2024tree,la2024can}, and Monte Carlo Tree Search (MCTS) \cite{li2024rethinkmcts,zhang2023planning,hu2024uncertainty,hao2023reasoning,feng2024alphazeroliketreesearchguidelarge} are used to explore the inherent reasoning process, often based on the self-play mechanism to reflect on previously generated contents to learn from itself \cite{haluptzok2022language,chen2023gaining,lu2023self,chen2023teaching,madaan2024self,shinn2024reflexion}.
During inference, error position can be beneficial in improving the reliability and performance of the model. With identification and analysis of where and why errors occur, recent research \cite{yao2024mulberry, luo2024improve, wu2025error} has made significant strides in quantifying and mitigating errors during model inference. Refinement \cite{madaan2024self, gou2023critic} and reflexion \cite{shinn2024reflexion, lee2025evolving} are also powerful techniques for enhancing the inference capabilities of LLMs, usually by enabling iterative improvement and self-correction.

\subsection{Model Composition}
Model composition technique gains notable attention in cross-tasks generalization. 
Traditional methods for multiple tasks are to train models on a mixture of datasets of different skills \cite{caruana1997multitask, chen2018gradnorm}, with the high cost of data mixing and lack of scalability of the model though. Model merging is a possible solution to this. Linear merging is a classic merging method that consists of simply averaging the model weights \cite{izmailov2018averaging, smith2017investigation}. Furthermore, Task Arithmetic \cite{ilharco2022editing} computes task vectors for each model, merges them linearly, and then adds back to the base, and SLERP \cite{white2016sampling} spherically interpolates the parameters of two models. Based on Task Arithmetic framework, TIES \cite{yadav2024ties} specifies the task vectors and applies a sign consensus algorithm to resolve interference between models, and DARE \cite{yu2024language} matches the performance of original models by random pruning.

Recently, LoRA merging methods are also widely applied to cross-task generalization. CAT \cite{prabhakar2024lora} introduces learnable linear concatenation of the LoRA layers, and Mixture of Experts(MoE) \cite{buehler2024x, feng2024mixture} method has input-dependent merging coefficients. Other linear merging methods of LoRAs, such as LoRA Hub \cite{huang2023lorahub}, involve additional cross-terms compared to simple concatenation. 

\section{Preliminaries}
\subsection{Monte-Carlo Tree Search}
Monte Carlo Tree Search (MCTS) is a decision-making algorithm widely used in environments with large state and action spaces, particularly in game AI and planning.  It incrementally builds search trees to estimate optimal actions by simulating random plays from various nodes and gradually improving action-value estimates based on simulation outcomes. Over iterations, this approach gradually converges to near-optimal decision-making policies. Notably, its integration with reinforcement learning has driven breakthroughs in systems like AlphaGo and AlphaZero \cite{silver2017mastering}, achieving superhuman performance in games.

Classical MCTS consists of four stages: selection, expansion, simulation, and backpropagation. It typically employs Upper Confidence Bounds for Trees (UCT) \cite{kocsis2006bandit}, which balances exploration and exploitation by guiding the search to promising nodes. After simulation, results propagate back through the tree, updating node values. However, MCTS struggles in domains with large action spaces, where excessive branching can degrade performance. Progressive Widening and Double Progressive Widening techniques have been proposed to mitigate this by dynamically limiting the number of actions considered at each decision node \cite{coulom2006efficient}.

\subsection{LoRA Finetuning}
LoRA (Low-Rank Adaptation) \cite{hu2021lora} fine-tuning is a technique used to adapt large pre-trained models, such as transformers, to specific tasks with minimal computational overhead. The key idea behind LoRA is to introduce low-rank matrices into the model's weight updates, which reduces the number of trainable parameters and makes fine-tuning more efficient. 

LoRA starts with a model that has been trained on a large dataset. During finetuning, instead of updating the full weight matrix $W \in \mathbb{R}^{m \times n}$, LoRA introduces two low-rank matrices $A \in \mathbb{R}^{m \times r}$ and $B \in \mathbb{R}^{r \times n}$, where $r \ll \min(m, n)$. The updated weight matrix $W'$ is then given by:
\begin{equation}
    W' = W + \Delta W = W + A \cdot B.
\end{equation}

During fine-tuning, only the matrices $A$ and $B$ are updated, while the original weight matrix $W$ remains frozen. This reduces the number of trainable parameters from $m \times n$ to $m \times r + r \times n$, which is much smaller when $r$ is small. For a given task with loss function $\mathcal{L}$, the objective is to minimize:
\begin{equation}
\mathcal{L}(y, f(x; W + A \cdot B)),
\end{equation}
where $y$ is the target output, $x$ is the input, and $f$ is the model's forward function.

By introducing low-rank matrices, both the number of trainable parameters and memory footprint are reduced. This approach is particularly useful in scenarios where computational resources are limited or when fine-tuning needs to be done quickly.

\section{Methodology}
In this section, we introduce the overall methodology of BDC, addressing challenges in the System2-to-System1 pipeline for code generation, specifically the complexity of hidden reasoning processes and heterogeneous data distributions. 
The proposed BDC pipeline consists of three main stages: 1) explore the System 2 knowledge via mutual verification and boosting between LLMs; 2) disentangle the obtained data into clusters over which composible LoRA experts are tuned; 3) customize problem solver by weighting over the composable LoRA experts using an input-aware hypernetwork.

\begin{figure*}
    \centering
    \includegraphics[width=1.0\linewidth]{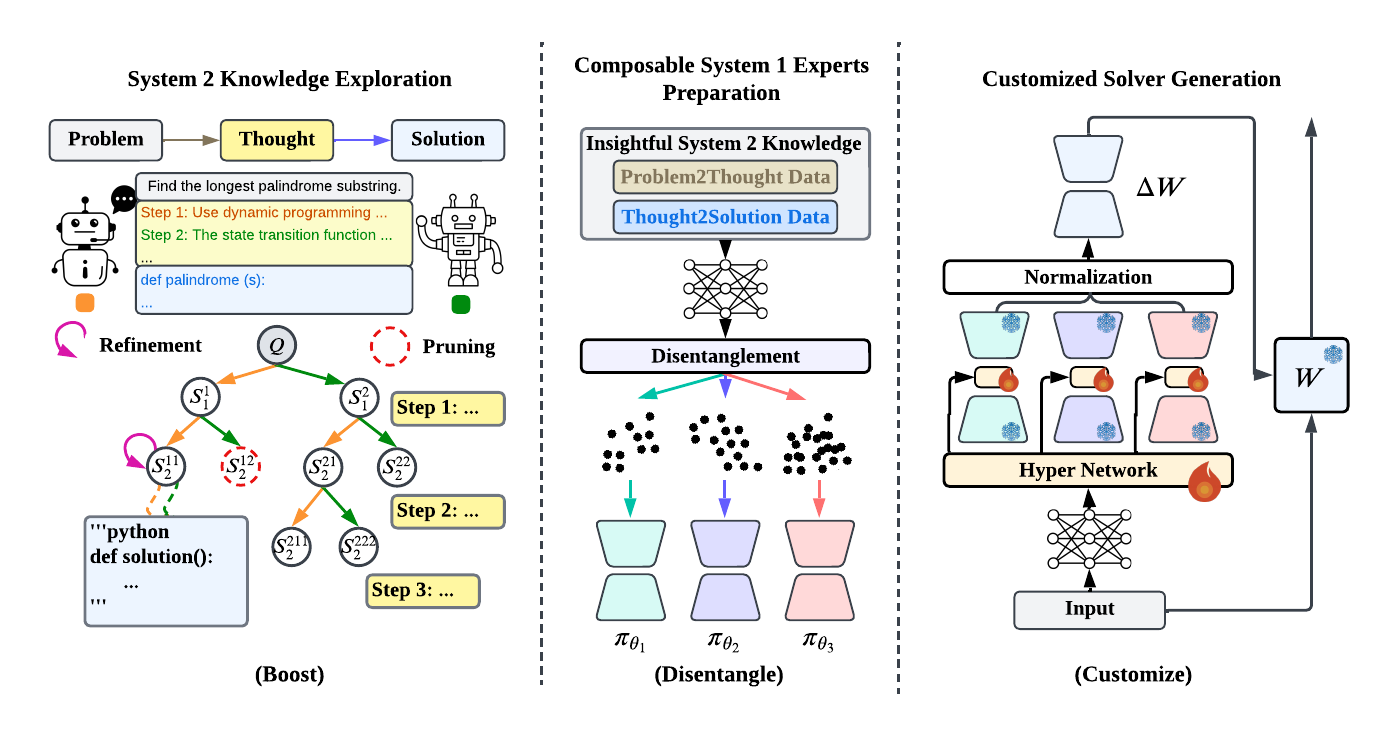}
   \caption{Illustration of the overall framework of \modelname.}
    \label{fig:overview}
\end{figure*}

\subsection{System 2 Knowledge Exploration}

In this subsection, we introduce the mechanism design for the data collection process. Due to the complex reasoning nature embodied, code blocks are hard to evaluate and estimate before mature. Reliable reward signals of a reasoning path therefore mainly depend on the dynamic compilation and execution feedbacks, which are extremely sparse and require extensive simulations. To simplify the generation paradigm and exploit the mutual verification capabilities of the collective searching, we decompose the generation process into two distinct stages: problem-to-thought and thought-to-solution.

\subsubsection{Problem-to-thought}

Traditional Monto-Carlo Tree Searching comprises three key operations in each iteration: (a) Select, (b) Expand, (c) Backup. In the problem-to-thought stage, we further extend MCTS by two distinct operations (d) Prune, and (e) Refine to reduce the searching space. We elaborate on these operations as follows.

\minisection{Select} 
Starting from the root, the reasoning path is prolonged by iteratively adding a specific child of the latest node. The operation is usually governed by certain policies, among which we adopt Probability-weighted Upper Confidence Bound(P-UCB) to balance the exploration and exploitation:
\begin{equation}
\mathrm{PUCB}(S_c)=Q(S)+c\cdot P(a|S_p)\cdot\frac{\sqrt{\log N(S)}}{1+N(S_c)},
\end{equation}
where $S_c$ is the state of the child node. $S$ and $Q(S)$ denote the parent node's state and value. $P(a|S)$ is the conditional probability of sampling the sequence $a$. $N(S)$ is the total number of times the parent node $S$ has been visited during simulations, while $N(S_c)$ tracks visits to the child node $S_c$. The selection process will stop if either a semantic or rule-based(e.g. length limits) terminal state encounters.

\minisection{Expand}
The Expand operation is triggered if a non-terminal leaf node of the tree is selected. A set of predefined LLM polices $\pi_0, \cdots, \pi_n$ generate subsequent thought sequences ${a_i}_n$ given the state $S$ of the current node, forming new leaf nodes:
\begin{equation}
    \forall i\in[n], P(a_i|S) \sim \pi_i(·|S).
\end{equation}


\minisection{Backup}
For well-defined problems, a reasoning path ${S_t}$ will eventually end at a terminal leaf node $S_T$ by iterating the Select and Expand operations. The reward $r_T$ is set according to the evaluation. We will skip the definition of reward $r_t$ and passrate $PR(S_t)$, which will be detailed in the explanation of the Simulate operation. The reward value is back-propagated along the reasoning path to update the state values of corresponding ancestor nodes:
\begin{equation}
    Q(S_{t-1}) = f(Q(S_t), r_t + \gamma PR(S_t)),
\end{equation}
where $f$ is the value function.

Additionally, the visit counts of ancestors are updated alongside the reasoning path:
\begin{equation}
    N(S_t) = N(S_t) + 1.
\end{equation}

\begin{figure}[h]
    \centering
    \includegraphics[width=1.0\linewidth]{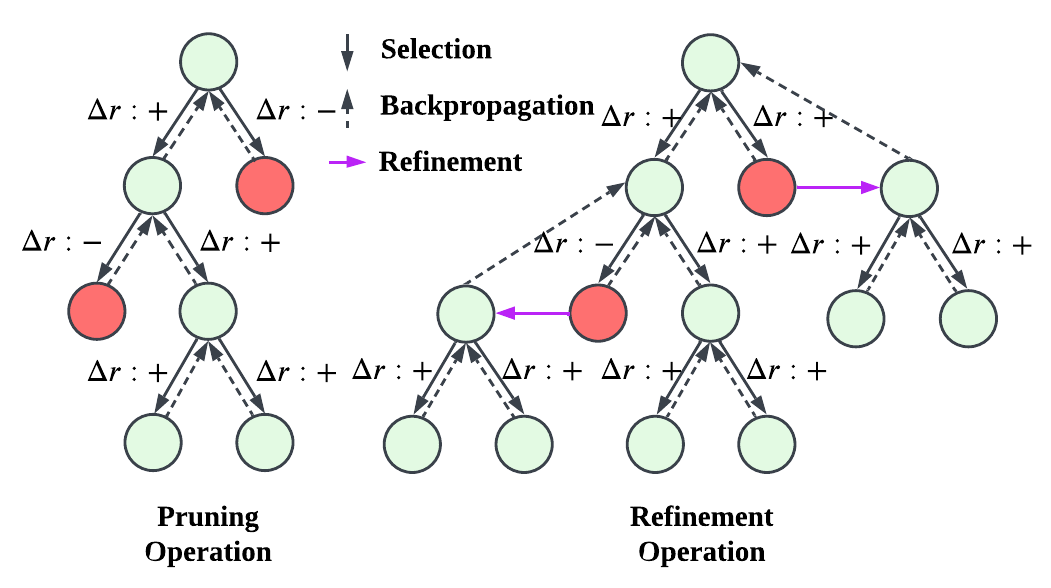}
    \caption{Pruning and refinement operations.}
    \label{fig:mc}
\end{figure}

We further extend and formalize reflective reason settings proposed in CoMCTS into Prune and Refine operations as shown in Figure~\ref{fig:mc}.

\minisection{Pruning}
The pruning operation on a selected node will examine and compare its passrate with that of its parent. With powerful LLMs, we consider valid and reasonable thoughts to bring non-negative influence solution seeking, thus featuring monotonically non-decreasing values in the passrate $PR(S_t) <= PR(S_{t+1})$.

A child node alleviating this principle will be considered as an ill node that introduces wrong thoughts. The ill node will be removed and trigger an instant Backup operation with zero reward to downweight its ancestors.

\minisection{Refine}
The truncated error and state information left by ill nodes will be analyzed in the Refine operations. To mitigate the bootstrapping bias of LLMs, a distinct policy LLM will be adopted to infer and summarize the information in natural language, which will be later utilized to refine and replace the ill nodes:
\begin{align}
    isIll(S^{\pi_i})&== 1,\nonumber\\
    Summary(S^{\pi_i}) &\sim \pi_{j}(Q(S^{\pi_i}),\\
    S^{\pi_i}, & BlockAnalysis(S^{\pi_i})),\nonumber
\end{align}
where $S^{\pi_i}$ denotes a ill node generated by $\pi_i$. A refined node is generated to replace the ill one:
\begin{equation}
    a' \sim \pi_{i}(Q(S^{\pi_i}), Summary).
\end{equation}

We enforce global and local constraints on possible times of calling Refine operation to avoid infinite loops and balance performance with compute budgets. A successful Refine operation will cause an in-place replacement of the ill-node, triggering another Backup operation to re-weight its ancestors.

\subsubsection{Thought-to-solution}

\minisection{Simulate}
For the thought-to-solution, we repurpose the Simulate operation for the collective solution generation process from the given state $S$. The operation will produce a set of possible solutions, each from a policy LLM:
\begin{equation}
    Solut.(S)_i \sim \pi_i(S).
\end{equation}

We define the passrate of a state as the average passrate of its corresponding solutions:
\begin{equation}
PR(S) = \frac{1}{n}\sum_{i}^{n} Passed(Solut.(S)_i),
\end{equation}
where $Passed(\cdot)$ represents the supervising signal from dynamic compilation and execution feedback.

The node's value $Q(S_t)$ is determined by its $PR(S_t)$ and reward $r_t$. Sincere additional solution string will be appended to a non-terminal state $S_t$ before evaluation, $PR(S_t)$ is an indirect supervising signal for the $S_t$, and the direct signal $r_t$ is set to zero.

The terminal state $S_T$ is treated as the unique solution itself since no string concatenation applies, therefore featuring a non-trivial reward $r_T$. Putting everything together, we have:
\begin{equation}
    Q(S) =\begin{cases} r_T &\text{if terminal,}\\\gamma PR(S_t) &\text{otherwise.}\end{cases}
\end{equation}


\subsection{System2-to-System1 Training}
\subsubsection{Heterogeneous Distribution Disentanglement}
\label{sec:dis}
After the data collection, the resulting training data obtained from the  MC-Tree-Of-Agents process consists of problem2thought data $D^{p2t}=\{\langle X_i^{p2t},y_i^{p2t}\rangle| i\in\mathbf{P}\}$ and thought2solution data $D^{t2s}=\{\langle X_i^{t2s},y_i^{t2s}\rangle| i\in\mathbf{P}\}$: $D_{train} = \{D^{p2t}, D^{t2s}\}$. As discussed in the introduction section, the latent patterns of coding problems are complex and tend to be heterogeneously distributed, e.g., the branching and recursion flow existing in the code blocks, different strategies of algorithm solutions, etc. Therefore, we disentangle the training data based on the latent semantics of the data into different clusters for fine-grained modeling.

The clustering objective can be summarized as below:
\begin{align}
    minimize_{\mathcal{C}} &\sum_k\sum_{i\in C_k} cosine(e_i, \mu_k), \\
    e_i = &\Phi_\theta (\langle X_i, y_i \rangle),\nonumber \\ 
    \mu_k = &mean\{e_i | i\in C_k\},\nonumber
\end{align}
where $\Phi_{theta}$ is the encoder of a code llm and $\mu_k$ denotes the centroid of cluster $C_k$.

\subsubsection{Composable LoRA Experts Preparation}
Having obtained the disentangled data clusters, we then finetune on them to obtain the meta LoRA experts for specialized experts of different aspects.
\begin{align}
    \forall C_k &\in \mathcal{C}, \nonumber\\
    \pi_{\theta_k} & \leftarrow SFT(\pi_{\theta}, \{\langle X_i, y_i\rangle | i \in C_k\}),
\end{align}
where $\pi_\theta$ denotes the base LLM and $\pi_{\theta_k}$ denotes the parameters of the LoRA adapter obtained by finetuning $\pi_\theta$ on $C_k$.

\subsubsection{Input-Aware Hypernetwork for Customized Solver}

Given specialized LoRA experts ${\pi_{\theta_1},\cdots,\pi_{\theta_K}}$ trained on distinct data clusters, we design an input-aware Hypernetwork $f(\cdot)$ to dynamically compose these experts through rank-wise adaption for customized problem solver.

For each input instance, the hypernetwork generates customized expert weights digesting its encoding and semantic distances to the cluster centroids. we identify "rank" as the minimal unit for aggregation and generate rank-wise weights for different experts at each decoding layer:
\begin{equation}
    G_i \leftarrow f(e_i, \langle cosine(e_i, \mu_1), \dots, cosine(e_i, \mu_K)\rangle),
\end{equation}
where $e_i$ is the encoding of input $X_i$, $G_i\in R^{K\times r\times 1}$ is the output weight matrix, $r$ is the rank of the LoRA matrices, and $K$ is the number of LoRA experts.

The aggregated $\Delta W$ of the linear projection layer is then obtained by
\begin{align}
    \mathbf{A}^* = [A_1, \dots, A_K] \odot G_i, \\
    \mathbf{\Delta W}^* = [B_1A_1^*, \dots, B_KA_K^*],\\
    \Delta W = ReduceSum(\mathbf{\Delta W}^*).
\end{align}


The projection output of $\Delta W$ is then merged during forwarding via:
\begin{equation}
    y= W_0x+\Delta Wx.
\end{equation}

We adopt a dedicated training phase for the Hypernetwork where all parameters are frozen except for the $f(\cdot)$. The training is supervised by the cross-entropy loss, with the randomly permuted input-output pairs from $D_{train}$.

\section{Experiments}
We conduct empirical studies starting from the following research questions.
\begin{itemize}[leftmargin=27pt]
    \item [\textbf{RQ1}] Does the proposed data collection algorithm explore insightful reasoning knowledge?
    \item [\textbf{RQ2}] Do the complex latent patterns of reasoning data impact the training performance?
    \item [\textbf{RQ3}] Can the disentangle-and-compose mechanism help to promote performance?
    \item [\textbf{RQ4}] Do the proposed input-aware hypernet work outperform other model composition techniques?
    \item [\textbf{RQ5}] How does DisenLoRA perform on untrained datasets?
\end{itemize}


\subsection{Setup}
In this section, we provide detailed setup information for the evaluation of the proposed \modelname, including datasets, trajectory data collection, and competing methods. 

The overall evaluation is conducted on two benchmark datasets: the competition-style APPS dataset and the CodeContest dataset. Both datasets categorize problems from easy to hard. We randomly sample problem subsets from each category of these two datasets. Each subset contains approximately 100 problems, except for the CodeContest-Hard category, which consists of around 50 problems due to inherent limitation in size. 

We conduct isolated assessments of both stages of \modelname to ensure a comprehensive comparison. 

\minisection{ Data collection} For Python code generation, we compare the performance of MCTS over different methods: zeroshot, LDB \cite{zhong2024ldb}, RAP \cite{hao2023reasoning}, Reflexion, LATS \cite{zhou2023language}, ToT and RethinkMCTS \cite{li2024rethinkmcts}. To mitigate the influence of factors such as context window limitations and instruction-following capabilities, we employ two advanced base models: GPT-4o-mini and Claude-3.5-Sonnet. Aligned with the purpose, we adopt a greedy decoding strategy for both models. Additionally, we provided peer comparisons between these two base models when driven by the MC-Tree-Of-Agents method in terms of their error position and refinement capability.

\minisection{ Fine-tuning} For fine-tuning, \modelname is compared against several alternative methods, including SFT on clustered subsets, TIES, DARES, and TWINS \cite{liu2023twins}.

\begin{table*}[ht]
\centering
\caption{Main results on System 2 knowledge exploration.}
\label{tab:datacollection}
\resizebox{0.9\textwidth}{!}{
\begin{tabular}{c|cccccc|cccc} 
\hline
& \multicolumn{6}{c|}{APPS} & \multicolumn{4}{c}{CodeContest}\\
       Models                 & \multicolumn{2}{c}{Intro.}                            & \multicolumn{2}{c}{Inter.} & \multicolumn{2}{c|}{Comp.} & \multicolumn{2}{c}{Easy} & \multicolumn{2}{c}{Hard} \\ \cline{2-11} 
                        & PR                     & AC                    & PR       & AC       & PR       & AC       & PR          & AC         & PR          & AC         \\ \hline
ZeroShot                & 56.56 & 35.00 &      40.57           &     19.00          &   23.67              &     9.00          &29.03& 19.61 &   28.24                 &19.23              \\
LDB                     & 60.64 & 40.00 &     46.78            &     22.00          &      21.00           &    8.00           & 34.76              & 25.58           & 33.52                   & 16.28                \\
RAP                     & 64.24                         & 39.00                         &   43.32              &     14.00          &         22.83        &    8.00           & 43.08              & 33.33            &39.99  &26.92                 \\
Reflexion               & 60.65                         & 40.00                         &  45.58               &     21.00          &      17.50           &     7.00          & 56.16              & 47.83           &34.09 &21.15                 \\
LATS                    & 69.46                         & 50.00                         &  45.86               &     20.00          &       21.83          &        7.00       &57.70              & 47.83           &39.10                    &30.77                 \\
ToT                     & 74.34                         & 55.00                         &  63.49               &     33.00          &         26.30        &       11.00        & 51.89              & 41.18           &49.07                    & 32.69                \\ 
RethinkMCTS                     & 76.60                         & 59.00                        &  74.35               &     49.00         &         42.50        &       28.00        & 60.84             & 51.53          & 55.79                & 48.04               \\ \hline
Single (GPT4omini) & 77.99                         & 60.00                         &     72.89            &      50.00            &      44.17         &      25.00     & 55.79     &48.04                &      45.72              & 26.92                \\
Single (Claude)    &  73.80  &  61.00   &  73.60   & 57.00 &      54.67           &      42.00         &                  58.75& 53.92                &68.41          &55.76             \\ \hline
MC-Tree-Of-Agents              & 79.72                         & 64.00                      &       79.42          &    63.00           &       59.17          &    45.00           & 62.49               &54.64             & 70.49                   &56.41                 \\ 
+ Pruning           & 85.18                         & 76.00                     & 81.95                & 67.00             &54.00              & 38.00            & 64.62            &  59.80         &73.12             &59.62                 \\
+ Refine      & 81.29                         & 68.00                    & 78.85              & 62.00            & 60.33             & 44.00            & 63.23             &  56.86          & 73.80                   &63.46\\                 \hline
\end{tabular}
}
\end{table*}

\begin{table*}[ht]
\centering
\caption{Main results on System2-to-System1 tuning.}
\label{tab:tuning}
\resizebox{\textwidth}{!}{
\begin{tabular}{ccccccccccccccc}
\hline
\multicolumn{15}{c}{ Meta-llama-3.1-instruct-8b}                                                                                                                                                                                                                  \\ \hline
\multicolumn{1}{c|}{}                                         & \multicolumn{2}{c}{Intro. (100)} & \multicolumn{2}{c}{Inter. (100)} & \multicolumn{2}{c|}{Comp. (100)}  & \multicolumn{2}{c|}{Overall}      & \multicolumn{2}{c}{Easy (102)} & \multicolumn{2}{c|}{Hard (51)}     & \multicolumn{2}{c}{Overall} \\ \cline{2-15} 
\multicolumn{1}{c|}{\multirow{-2}{*}{Finetune Method}}        & PR              & AC             & PR              & AC             & PR    & \multicolumn{1}{c|}{AC}   & PR    & \multicolumn{1}{c|}{AC}   & PR             & AC            & PR    & \multicolumn{1}{c|}{AC}    & PR           & AC           \\ \hline
\multicolumn{1}{c|}{w/o tuning}                               & 21.14           & 4.00           & 20.72           & 4.00           & 12.83 & \multicolumn{1}{c|}{1.00} & 18.23 & \multicolumn{1}{c|}{3.00} & 25.54          & 17.65         & 15.46 & \multicolumn{1}{c|}{5.77}  & 22.18        & 13.69        \\ \hline
\multicolumn{1}{c|}{SFT on all}                               & 22.55           & 7.00           & 26.40           & 3.00           & 10.67 & \multicolumn{1}{c|}{1.00} & 19.87 & \multicolumn{1}{c|}{3.67} & 25.33          & 17.65         & 16.73 & \multicolumn{1}{c|}{7.69}  & 22.46        & 14.33        \\ \hline
\multicolumn{1}{c|}{ SFT on cluster 0} & 20.67           & 6.00           & 24.23           & 3.00           & 11.50 & \multicolumn{1}{c|}{1.00} & 18.80 & \multicolumn{1}{c|}{3.33} & 27.31          & 17.65         & 11.69 & \multicolumn{1}{c|}{1.92}  & 22.10        & 12.41        \\
\multicolumn{1}{c|}{ SFT on cluster 1} & 21.22           & 4.00           & 20.69           & 4.00           & 12.00 & \multicolumn{1}{c|}{2.00} & 17.97 & \multicolumn{1}{c|}{3.33} & 27.78          & 20.59         & 18.12 & \multicolumn{1}{c|}{9.62}  & 24.56        & 16.93        \\
\multicolumn{1}{c|}{SFT on cluster 2}                         & 16.65           & 7.00           & 23.97           & 3.00           & 17.33 & \multicolumn{1}{c|}{4.00} & 19.32 & \multicolumn{1}{c|}{4.67} & 26.82          & 20.59         & 19.50 & \multicolumn{1}{c|}{9.62}  & 24.38        & 16.93        \\ \hline
\multicolumn{1}{c|}{Ties}                                     & 22.75           & 4.00           & 23.06           & 4.00           & 12.67 & \multicolumn{1}{c|}{4.00} & 19.49 & \multicolumn{1}{c|}{4.00} & 26.64          & 21.57         & 18.71 & \multicolumn{1}{c|}{9.62}  & 24.00        & 17.59        \\
\multicolumn{1}{c|}{Dare}                                     & 24.97           & 7.00           & 26.66           & 5.00           & 12.50 & \multicolumn{1}{c|}{3.00} & 21.38 & \multicolumn{1}{c|}{5.00} & 23.05          & 13.73         & 19.65 & \multicolumn{1}{c|}{15.38} & 21.92        & 14.28        \\
\multicolumn{1}{c|}{Twin}                                     & 19.10           & 5.00           & 23.85           & 5.00           & 8.50  & \multicolumn{1}{c|}{1.00} & 17.15 & \multicolumn{1}{c|}{3.67} & 26.87          & 17.64         & 12.92 & \multicolumn{1}{c|}{9.62}  & 22.22        & 14.97        \\ \hline
\multicolumn{1}{c|}{DisenLoRA}                & \textbf{27.11}           & \textbf{9.00}           & 23.11           & 3.00           & 11.50 & \multicolumn{1}{c|}{\textbf{4.00}} & \textbf{20.57} & \multicolumn{1}{c|}{\textbf{5.33}} & \textbf{32.24}          & \textbf{22.55}         & 19.43 & \multicolumn{1}{c|}{9.62}  & \textbf{27.97}        & \textbf{18.24}        \\ \hline
\end{tabular}
}
\end{table*}

\subsection{Empirical Analysis and Discussion}
\subsubsection{\textbf{RQ1}. MC-Tree-Of-Agents}

We evaluate MC-Tree-Of-Agents against widely-used baseline methods, the results are summarized in Table~\ref{tab:datacollection}.
From the results, we can draw the following conclusions. 
\begin{itemize}[leftmargin=10pt]
    \item The proposed MC-Tree-Of-Agents outperforms all the baseline methods, which effectively explores the insightful  System 2 knowledge. 
    \item Comparing with the single LLM as agents version, MC-Tree-Of-Agents allows for mutual verification and boosting between different LLMs, offering a superior performance over each distinct-LLM-as-agent method. This showcases the effectiveness of the interaction between LLMs of different wisdom.
    \item The pruning and refinement operations both contribute to the final performance, offering a notable accuracy gain. This validates that the designed pruning and refinement mechanism, based on the difference between rewards of parent-child nodes, can save the algorithm from erroneous exploration and lead to beneficial directions in limited rollouts.
\end{itemize}

\subsubsection{\textbf{RQ2}. Impact of latent patterns}

\begin{figure}[h!]
    \centering
    \includegraphics[width=0.8\linewidth]{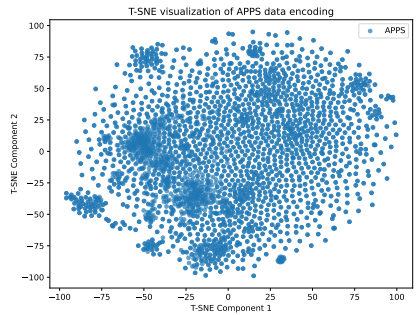}
    \caption{T-sne visualization of the APPS data encoding.}
    \vspace{-10pt}
    \label{fig:t-sne}
\end{figure}

To study the distribution of the latent patterns of coding problems, we first conduct the T-SNE visualization on the encodings of reasoning data collected by MC-Tree-Of-Agents on APPS dataset. The visualization is displayed in Figure~\ref{fig:t-sne}.

From the visualization, we can see that there different clusters of data distributions existing in the latent reasoning semantic space, which poses a potential challenge to robust and generalizable LLMs on code.

Furthermore, we perform finetuning on different clusters of data obtained in Section~\ref{sec:dis} and evaluate the corresponding models on the test data. From the results in Table~\ref{tab:tuning}, we can see the following conclusions. 1) LLMs finetuning on all the clusters can offer better performance than that of the non-tuning version, validating the quality of the collected System2 knowledge data. 2) Llm experts obtained from different clusters show different performances on different levels of tasks. One expert can demonstrate outstanding capability on one level of tasks, even outperforming the LLM finetuning on all the data, while performing weakly on a different level of task. This phenomenon further justifies the heterogeneous latent patterns of data distribution and serves as supportive evidence for disentangling LLM experts.

\subsubsection{\textbf{RQ3}. Effectiveness of the Experts Composition}

During the empirical study, we test different model merging methods that combine wisdom from different LoRA experts. We evaluate the well-known Ties, Dare, and the recently proposed TWIN merging methods. All of them yield a static composed model that takes in the strength of the candidate experts to be merged via solving parameter interference. From the results, we can see that merging over decomposed LoRA-experts can offer more robust problem solvers, outperforming the simple train-once-for-all mechanism. The experiments justify our major rationale that disentanglement-and-compose pipeline can offer more robust System2-to-System1 performance.

\subsubsection{\textbf{RQ4}. Superiority of DisenLoRA over other composition methods}

Although the static-composed expert model can promote robustness to some extent, its static nature lacks flexibility to different styles of inputs. As discussed in the previous contents, the data distribution of coding problems is complex, making the one-fits-all mechanism easy to fail. Therefore, we design DisenLoRA algorithm to yield a customized problem solver with input-awareness. From the results, we can see that  DisenLoRA outperforms the competing merging methods, validating the effectiveness of the proposed input-aware hypernetwork that dynamically aggregates the candidate composable LoRA experts at a rank-wise level.

\subsubsection{\textbf{RQ5}. Discussion of the Cross-Dataset Generalization of DisenLoRA}

Despite the flexibility offered by the input-aware hypernetwork, its performance may degrade on new datasets where the hypernetwork is not trained. To study this scenario, we use the model trained on APPS to generate solutions for CodeContest and use the model trained on CodeContest to generate solutions for APPS. The results are displayed in Table~\ref{tab:ood4code}.

\begin{table}[h]
    \centering
    \caption{Cross-dataset generalization test.}
    \label{tab:ood4code}
    \resizebox{0.45\textwidth}{!}{
    \begin{tabular}{c|cc|cc}
    \hline
        OOD Dataset & \multicolumn{2}{c|}{APPS}&\multicolumn{2}{c}{CodeContest}\\\hline
      Method   & PR & AC &PR & AC\\
      \hline
      w/o tuning   & 18.23 &	3.00 & 22.18 &	13.69 \\
      w/ SFT & 17.44 &	4.33 & 20.99	&14.29 \\
      \hline
      DisenLoRA & 18.25 &	4.33& 25.09&14.34\\ \hline
    \end{tabular}
    }
\end{table}

From the results, we can see that the proposed DisenLoRA has the generalization ability to the untrained dataset, outperforming the train-once-for-all mechanism still. This demonstrates that the parameters of the trained hypernetwork have the awareness of semantic similarities across datasets.

\section{Conclusion}
We identify the complexity of inherent reasoning exploration and the heterogeneous data distribution problems that hinder the performance of System2-to-System1 methods. Correspondingly, we propose the BDC pipeline that explores insightful System2 knowledge via mutually \textbf{B}oosting between llm agents, \textbf{D}isentangle heterogeneous data distribution for composable LoRA experts, and \textbf{C}ustomize problem solver for each instance, offering flexibility and robustness.
Correspondingly, we propose the MC-Tree-Of-Agents algorithm to efficiently and effectively explore the insightful System2 knowledge via mutual verification and boosting of different LLM agents, armed with reward-guided pruning and refinement to explore more beneficial states in limited rollouts for better performance. 
Additionally, we design an input-aware hypernetwork to aggregate over the disentangled composable LoRA experts trained on different clusters of data collected from MC-Tree-Of-Agents. This mechanism offers a customized problem solver for each data instance.
Various experiments and discussions validate the effectiveness of different model components.

\section*{Limitations}
While our work presents an efficient pipeline for transferring specialized knowledge from collective system-2-like LLMs to locally deployed language models through multiple LoRA adapters—enabling rapid, precise, system-1-like reasoning—three limitations merit discussion. First, despite code generation serving as an effective proxy for complex reasoning, our evaluation is restricted to this domain, leaving open questions about generalizability to broader textual reasoning tasks (e.g., commonsense reasoning and semantic parsing). Second, while we focus on their performance on the specific benchmarks, the safety alignment of derived models remains unaddressed. Systematic evaluation is required to assess whether our distilled experts preserve human values and mitigate harmful outputs. Finally, our ensemble methodology for LoRA experts, while input-aware, does not fully exploit potential sparsity optimizations in parameter activation, leaving room for computational efficiency improvements through advanced routing mechanisms.

\section*{Acknowledgments}

This document has been adapted by Emily Allaway from the instructions for earlier ACL and NAACL proceedings, including those for NAACL 2024 by Steven Bethard, Ryan Cotterell and Rui Yan,
ACL 2019 by Douwe Kiela and Ivan Vuli\'{c},
NAACL 2019 by Stephanie Lukin and Alla Roskovskaya,
ACL 2018 by Shay Cohen, Kevin Gimpel, and Wei Lu,
NAACL 2018 by Margaret Mitchell and Stephanie Lukin,
Bib\TeX{} suggestions for (NA)ACL 2017/2018 from Jason Eisner,
ACL 2017 by Dan Gildea and Min-Yen Kan,
NAACL 2017 by Margaret Mitchell,
ACL 2012 by Maggie Li and Michael White,
ACL 2010 by Jing-Shin Chang and Philipp Koehn,
ACL 2008 by Johanna D. Moore, Simone Teufel, James Allan, and Sadaoki Furui,
ACL 2005 by Hwee Tou Ng and Kemal Oflazer,
ACL 2002 by Eugene Charniak and Dekang Lin,
and earlier ACL and EACL formats written by several people, including
John Chen, Henry S. Thompson and Donald Walker.
Additional elements were taken from the formatting instructions of the \emph{International Joint Conference on Artificial Intelligence} and the \emph{Conference on Computer Vision and Pattern Recognition}.

\bibliography{custom}

\appendix

\end{document}